\documentclass{article}
\usepackage[utf8]{inputenc}
\usepackage{spconf,graphicx}
\usepackage{amsmath,amssymb,amsfonts}
\usepackage{amssymb}
\usepackage[dvipsnames]{xcolor}
\usepackage{mathtools}
\usepackage{multirow}
\usepackage{lipsum}
\usepackage{subcaption}

\usepackage[normalem]{ulem}
\usepackage{color,soul}
\usepackage{booktabs,makecell}
\usepackage{tabularx}

\usepackage{algorithm}
\usepackage{algpseudocode}

\usepackage{array, booktabs}
\usepackage{hyperref}
\newcolumntype{C}[1]{>{\centering\arraybackslash}p{#1}}

\def\BibTeX{{\rm B\kern-.05em{\sc i\kern-.025em b}\kern-.08em
    T\kern-.1667em\lower.7ex\hbox{E}\kern-.125emX}}

\usepackage{array}
\newcolumntype{P}[1]{>{\centering\arraybackslash}p{#1}}
\newcolumntype{L}[1]{>{\raggedright\arraybackslash}p{#1}}

\begin{document}

\title{ChromouVQA: Benchmarking Vision-Language Models under Chromatic Camouflaged Images}

\name{
    \begin{tabular}{@{}c@{}}
        Yunfei Zhang\textsuperscript{1,*},
        Yizhuo He\textsuperscript{2,*},
        Yuanxun Shao\textsuperscript{3},
        Zhengtao Yao\textsuperscript{4}, \\
        Haoyan Xu\textsuperscript{4},
        Junhao Dong\textsuperscript{5},
        Zhen Yao\textsuperscript{6},
        Zhikang Dong\textsuperscript{7,$\dagger$}
    \end{tabular}
    \thanks{$^{*}$Equal contribution.}
    \thanks{$^{\dagger}$Project Leader.}
}

\address{
$^{1}$Amazon \quad
$^{2}$Google \quad
$^{3}$MurcuryMind \quad
$^{4}$University of Southern California \\
$^{5}$Nanyang Technological University \quad
$^{6}$Lehigh University \quad
$^{7}$Stony Brook University
}

\maketitle
\begin{abstract}
Vision-Language Models (VLMs) have advanced multimodal understanding, yet still struggle when targets are embedded in cluttered backgrounds requiring figure–ground segregation. To address this, we introduce ChromouVQA, a large-scale, multi-task benchmark based on Ishihara-style chromatic camouflaged images. We extend classic dot plates with multiple fill geometries and vary chromatic separation, density, size, occlusion, and rotation, recording full metadata for reproducibility. The benchmark covers nine vision-question-answering tasks, including recognition, counting, comparison, and spatial reasoning. Evaluations of humans and VLMs reveal large gaps, especially under subtle chromatic contrast or disruptive geometric fills. We also propose a model-agnostic contrastive recipe aligning silhouettes with their camouflaged renderings, improving recovery of global shapes. ChromouVQA provides a compact, controlled benchmark for reproducible evaluation and extension. Code and dataset are available at \url{https://github.com/Chromou-VQA-Benchmark/Chromou-VQA}.

\end{abstract}

\begin{keywords}
Camouflage Perception, Vision--Language Models, Visual Question Answering, Benchmark
\end{keywords}

\section{Introduction}
\label{sec:intro}

Vision–Language Models (VLMs) have made rapid progress, demonstrating impressive performance across a wide range of multimodal tasks such as image and video captioning \cite{sharma2018conceptual,beedu2025mamba}, visual question answering \cite{antol2015vqa,liu2024tackling}, and open-domain reasoning \cite{bai2025qwen2,dong2024musechat}. In addition, the broader literature on language and multimodal
modeling continues to expand across diverse tasks \cite{fu2024detecting,dong2023cp,xin2024let,dong2024mapping,xin2025improving,dong2024cp,wang2025digital,dong2025every,dong2025artificial,yao2025generative}. On many established benchmarks, modern models even approach or surpass human-level accuracy.

\begin{figure*}[!h]
  \centering
   \includegraphics[width=\linewidth]{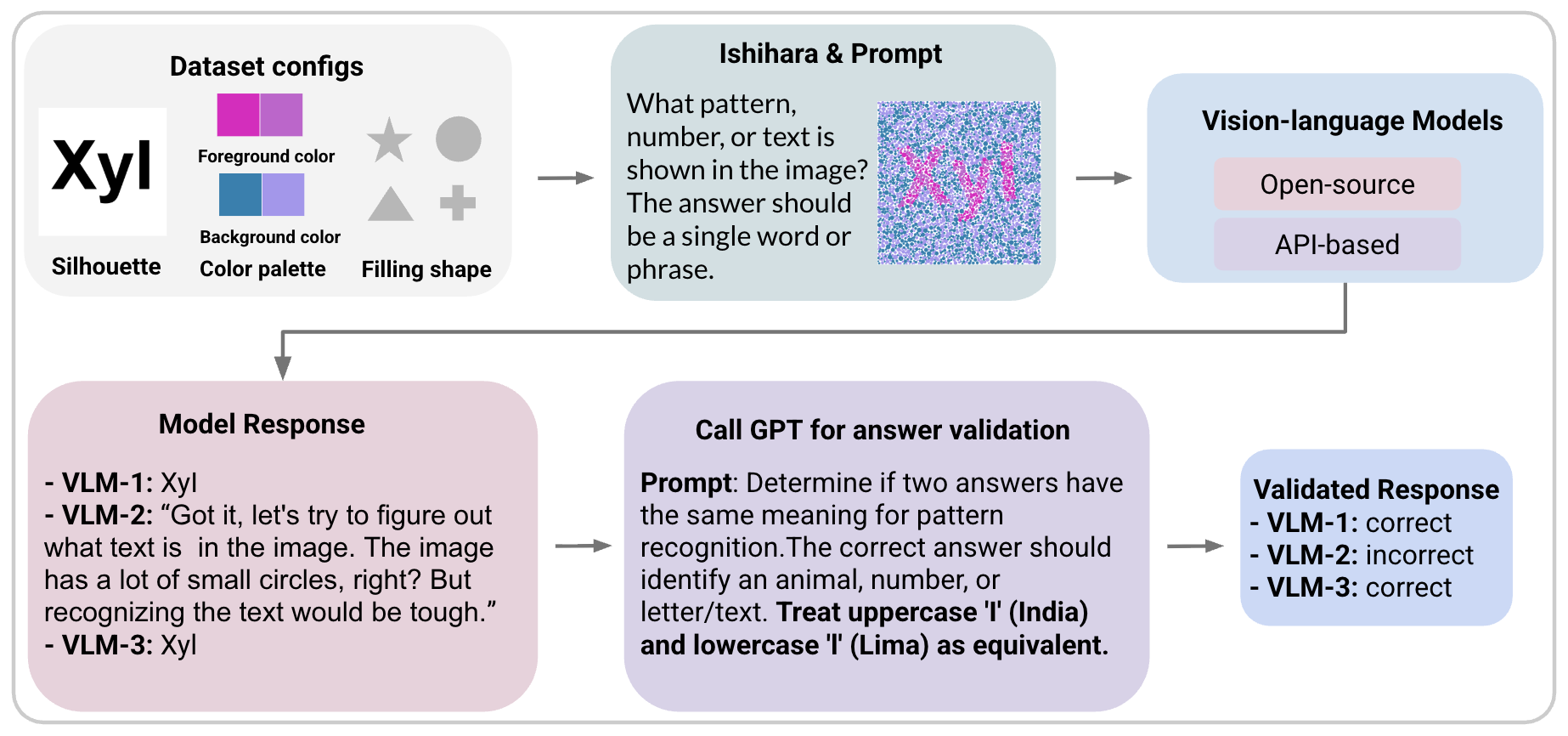}
   \caption{Dataset generation and model inference pipeline (rotation-invariant perception task).}
   \label{fig:flow_chart_resized}
\end{figure*}

Most existing benchmarks emphasize clean, high-contrast images where objects are easily distinguishable from their surroundings. Yet visual information can remain accessible to humans but challenging for models when foreground and background share similar colors and textures, causing local cues to fail. For example, safety-critical settings such as aviation use cockpit displays and warning icons that may appear against patterned or color-similar backdrops; humans can still recognize these signals once the figure emerges, but current VLMs often misidentify or overlook them. This gap raises a central question: can VLMs reliably perceive and ground content when chromatic camouflage disrupts local cues?

We introduce \emph{ChromouVQA}, a benchmark for systematic evaluation of VLMs under chromatic camouflage. We draw inspiration from Ishihara-style designs, not for their medical use in color-vision testing, but because they offer several properties that make them particularly well-suited for controlled benchmarking. First, camouflage difficulty can be precisely tuned by adjusting color distance, dot density, and pattern geometry, allowing interpretable performance curves. Second, recognition requires integrating subtle chromatic differences across the entire image, which stresses models that rely heavily on local patches rather than global structure. Third, the synthetic generator enables scalability and reproducibility, producing large-scale data with complete metadata while avoiding confounds from uncontrolled natural imagery such as lighting or clutter. Finally, Ishihara-style patterns offer a consistent human reference, serving as a strong baseline.

The dataset contains 70,200 camouflaged renderings derived from 17,100 silhouettes, spanning 61 palette–shape configurations and paired with nine question–answer tasks. These tasks extend beyond recognition to include reasoning skills such as counting, comparison, and arithmetic, all under camouflage conditions. Human baselines show near-ceiling accuracy, while state-of-the-art VLMs struggle, with gaps that widen under subtle chromatic contrast and non-dot fills. To mitigate this, we propose a contrastive adaptation framework in which clean silhouettes act as anchors and negatives, encouraging models to recover hidden patterns. This improves performance across different backbones and establishes a reference baseline for future study.

Our contributions are as follows:
(1) We introduce ChromouVQA, a large-scale, multi-task dataset of Ishihara-style camouflaged images for vision-language tasks, together with a novel, highly flexible and controllable image generation pipeline. (2) We conduct a comprehensive empirical study across a wide range of open-source and proprietary VLMs, highlighting significant performance gaps between current models and human perception on camouflaged image tasks.
(3) We propose a contrastive training framework that leverages silhouette–camouflage pairs to enhance global shape recovery, providing a strong baseline for this domain.

\section{Related Work}
\label{sec:related_word}

\textbf{Vision–Language Models.} Vision–Language Models (VLMs) have advanced rapidly in recent years. LLaVA \cite{li2024llava} demonstrates the effectiveness of instruction tuning, while Qwen2.5-VL \cite{bai2025qwen2} introduces dynamic resolution. Other models, such as InternVL \cite{chen2024internvl}, LLaMA 3 \cite{grattafiori2024llama}, and Video-LLaMA \cite{zhang2025videollama}, rely on large-scale pretraining across diverse modalities. Proprietary systems including GPT-4V \cite{openai2023gpt4} and Gemini \cite{google2024gemini} report near-human performance on vision tasks. Despite these advances, existing models have not been systematically evaluated on their ability to extract abstract patterns in complex or noisy settings—particularly in chromatic camouflage images.

\noindent \textbf{Visual Question Answering.} Visual Question Answering (VQA) is a rapidly developing research area in which models generate text answers to questions grounded in images. Early work \cite{antol2015vqa} introduced a large-scale benchmark that established VQA as a core multimodal task. Subsequent datasets expanded the scope: OK-VQA \cite{marino2019ok} emphasizes knowledge-based reasoning requiring external information; VAGUE \cite{nam2025vaguevisualcontextsclarify} evaluates models’ ability to resolve ambiguous queries through visual context; and PMC-VQA \cite{zhang2023pmc} provides diverse medical images with curated question–answer pairs across multiple modalities and diseases. However, no existing VQA benchmark comprehensively evaluates models’ ability to interpret chromatic camouflage images.

\noindent \textbf{Vision Illusion and Hallucination}. Visual illusion and hallucination remain key challenges for VLMs. SpookyBench \cite{upadhyay2025timeblindnessvideolanguagemodels} evaluates temporal pattern recognition when spatial cues are obscured. IllusionVQA \cite{shahgir2024illusionvqachallengingopticalillusion} provides optical illusions and ambiguous scenes for testing comprehension and localization. HallusionBench \cite{Hallusionbench} analyzes image-context reasoning through expert-designed questions, highlighting failures such as hallucination. \cite{zhang-etal-2023-grounding,ullman2024illusionillusionvisionlanguagemodels} build illusion datasets showing VLMs often misidentify normal images as illusions. RCID \cite{mao2024evaluatingmodelperceptioncolor} further demonstrates that models confuse pixel values with perceptual cues.

\section{Dataset Generation}
\label{sec:dataset_generation}

\begin{table*}[htbp]
\centering
\small
\resizebox{\textwidth}{!}{
\begin{tabular}{lccccccccccc}
\toprule
\textbf{} 
& \textbf{Silhouette} 
& \textbf{Math} 
& \textbf{Recognition} 
& \textbf{Occlusion} 
& \textbf{Rotation} 
& \textbf{Count} 
& \textbf{Enumeration} 
& \textbf{Spot\newline Difference} 
& \textbf{Size\newline Comparison} 
& \textbf{Size\newline Sort} 
& \textbf{Overall} \\
\midrule
Humans (average) & 100\% & 84.6\% & 89.7\% & 82.1\% & 81.6\% & 97.4\% & 87.1\% & 87.5\% & 100\% & 85.4\% & 88.4\% \\
\midrule
GPT-4o               & 77.4\% & \textbf{13.1\%} & \textbf{19.0\%} & 11.4\% & \textbf{11.6\%} & 19.2\% & \textbf{3.3\%} & \textbf{36.1\%} & 53.9\% & \textbf{16.3\%} & \textbf{20.4\%} \\
GPT-4o Mini          & 52.2\% & 11.9\% & 17.7\% & \textbf{11.7\%} & 11.5\% & 28.5\% & 2.2\% & 32.0\% & 17.8\% & 1.2\% & 14.9\% \\
Claude Sonnet 4         & \textbf{86.7\%} & 0.3\% & 1.7\% & 1.5\% & 0.9\% & \textbf{52.8\%} & 0.2\% & 31.3\% & \textbf{59.3\%} & 12.1\% & 18.1\% \\
Claude Haiku 3.5        & 72.0\% & 0.5\% & 1.5\% & 3.0\% & 1.7\% & 42.1\% & 0.1\% & 1.1\% & 28.4\% & 2.5\% & 9.2\% \\
\midrule
GLM4.1v-Thinking -9B\cite{vteam2025glm41vthinkingversatilemultimodalreasoning}              & 27.5\% & 1.0\% & \textbf{8.0\%} & \textbf{4.9\%} & \textbf{5.5\%} & 0.0\% & 0.6\% & 43.0\% & 35.1\% & 0.0\% & 10.9\% \\
InternVL-2.5-8B\cite{chen2025expandingperformanceboundariesopensource}        & 56.8\% & \textbf{2.3\%} & 6.9\% & 4.3\% & 3.8\% & \textbf{25.8\%} & 1.7\% & \textbf{44.3\%} & \textbf{68.7\%}      & \textbf{9.4\%} & \textbf{18.6\%} \\
LLaVA-1.6-7B \cite{li2024llava2}           & 33.1\% & 0.7\% & 4.8\% & 3.3\% & 2.9\% & 14.5\% & 0.2\% & 30.1\% & 16.2\% & 0.0\% & 8.1\% \\
Llama4-Scout               & 49.2\% & 0.2\% & 3.3\% & 2.4\% & 2.8\% & 1.1\% & \textbf{3.2\%} & 33.6\% & 32.8\% & 0.8\% & 8.9\% \\
Llama3.2-Vision-11B \cite{grattafiori2024llama}               & 42.3\% & 0.5\% & 0.6\% & 0.5\% & 0.4\% & 4.6\% & 0.3\% & 25.0\% & 26.8\% & 0.7\% & 6.6\% \\
Phi3-Vision-4.2B \cite{abdin2024phi3}               & 35.6\% & 0.3\% & 2.9\% & 2.5\% & 1.7\% & 6.9\% & 0.1\% & 25.7\% & 27.2\% & 0.0\% & 7.5\% \\
Phi4-Multimodal-5.6B \cite{microsoft2025phi4minitechnicalreportcompact}          & 12.0\% & 0.4\% & 0.3\% & 0.2\% & 0.2\% & 8.3\% & 0.0\% & 22.5\% & 4.9\% & 0.6\% & 4.2\% \\
Qwen2.5-Omni-7B \cite{xu2025qwen25omnitechnicalreport}         & 57.6\% & 0.9\% & 0.7\% & 0.5\% & 0.7\% & 2.4\% & 0.2\% & 20.6\% & 12.5\% & 0.1\% & 4.3\% \\
Qwen2.5-VL-3B \cite{bai2025qwen2}        & 49.9\% & 1.2\% & 0.3\% & 0.3\% & 0.2\% & 0.0\% & 0.1\% & 24.3\% & 25.4\% & 2.0\% & 6.0\% \\
Qwen2.5-VL-7B \cite{bai2025qwen2}        & \textbf{60.4\%} & 1.1\% & 0.1\% & 0.2\% & 0.2\% & 0.8\% & 0.1\% & 27.5\% & 26.3\% & 6.0\% & 6.9\% \\
Qwen2.5-VL-32B \cite{bai2025qwen2}       & 39.2\% & 1.2\% & 0.7\% & 0.8\% & 0.5\% & 2.9\% & 2.5\% & 5.1\% & 4.9\% & 0.3\% & 2.1\% \\
\midrule
Contra-InternVL         & 68.5\% & 52.1\% & 44.3\% & 29.5\% & 27.8\% & \textbf{88.2\%} & 27.0\% & \textbf{89.7\%} & \textbf{92.5\%} & \textbf{94.3\%} & 60.6\% \\
Contra-LLaVA            & 65.2\% & 46.8\% & 40.1\% & \textbf{31.7\%} & 26.9\% & 84.5\% & 25.3\% & 85.4\% & 83.1\% & 86.0\% & 56.6\% \\
Contra-Qwen            & \textbf{71.9\%} & \textbf{59.6\%} & \textbf{47.7\%} & 30.0\% & \textbf{28.0\%} & 87.9\% & \textbf{28.7\%} & 84.9\% & 89.8\% & 92.7\% & \textbf{61.0\%} \\
\bottomrule
\end{tabular}
}
\caption{Task-wise Evaluation Accuracy on the Complete Color Dataset. Bolded values mark the highest accuracy for each model within its respective category. \textit{Overall} is the mean accuracy across the nine tasks, excluding Silhouette.}

\label{tab:recognition_accuracy_detailed_weighted}
\end{table*}

\begin{figure*}[!h]
  \centering
   \includegraphics[width=\linewidth]{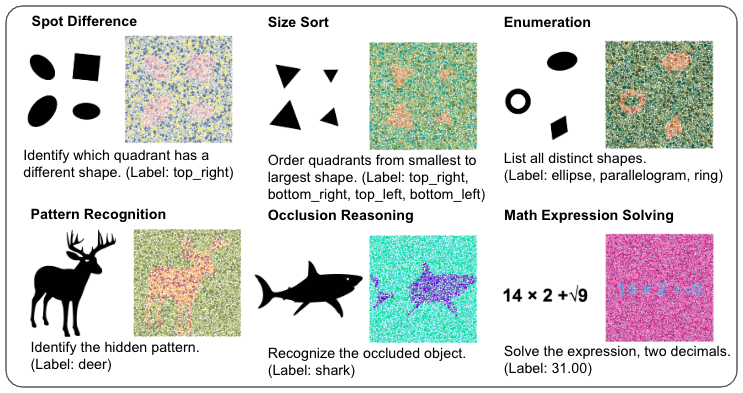}
   \caption{Example tasks in ChromouVQA, details are in the Dataset Generation section.}
   \label{fig:example_tasks}
\end{figure*}

Inspired by \cite{franciscouzo2025ishihara}, we propose a highly controllable image generation pipeline that supports different QA tasks, color settings, and filling shapes.

\subsection{Unified Task Suite: Visual Reasoning and Recognition}
We introduce nine tasks that increase visual complexity and systematically scale across content types, evaluating both basic visual discrimination and advanced reasoning under challenging conditions. (1) \textbf{Count}. Report the number of shapes constructed by filling shapes. (2) \textbf{Enumeration}. List all distinct shapes, sorted alphabetically. (3) \textbf{Spot Difference}. Identify the quadrant with the different shape. (4) \textbf{Size Comparison}. Identify the quadrant with the largest shape. (5) \textbf{Size Sort}. Order quadrants by shape size, smallest to largest. (6) \textbf{Pattern Recognition}. Recognize text, a 3-digit number, or one of the 134 animal silhouettes. (7) \textbf{Rotation-Invariant Perception}. Identify rotated content. (8) \textbf{Occlusion Reasoning}. Recognize content partially hidden by circles. (9) \textbf{Mathematical Calculation}. Solve the expressions shown.

\subsection{Chromatic Separation Control}
\label{subsec:vlm}

We control foreground–background color separation with two palette sources. First, nine palettes are extracted from real Ishihara plates, spanning Dual-color (two per side), Tri-color (three–four), and Multi-color (over four). Second, to increase diversity while ensuring sRGB compatibility, we sample 2–5 foreground and background colors per plate, yielding 16 configurations. Colors are chosen to avoid near-duplicates and maintain moderate hue–brightness separation, following principles of color specification~\cite{wyszecki1982color}.

\subsection{Diverse Shapes Filling}

\begin{figure}[!h]
  \centering
  \begin{subfigure}[b]{0.49\linewidth}
    \centering
    \includegraphics[width=\linewidth]{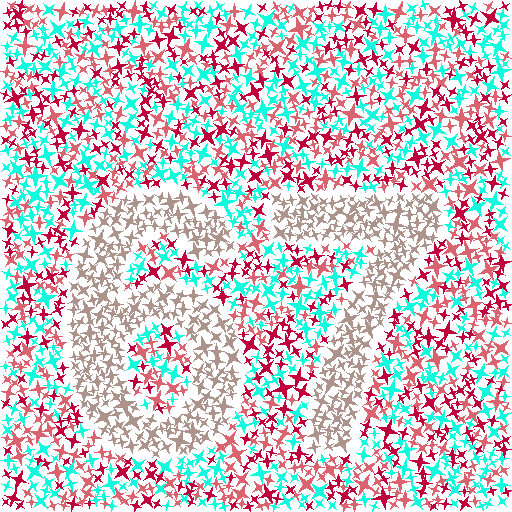}
  \end{subfigure}
  \hfill
  \begin{subfigure}[b]{0.49\linewidth}
    \centering
    \includegraphics[width=\linewidth]{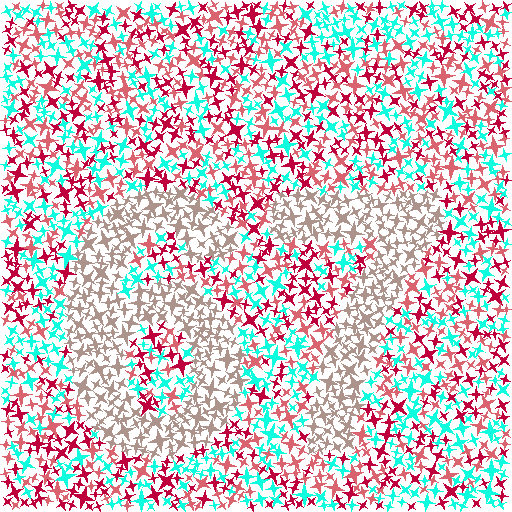}
  \end{subfigure}
  \caption{Filling algorithm from \cite{franciscouzo2025ishihara} (left) and our ray casting-based filling method (right).}
  \label{fig:algo_comparison}
\end{figure}

We extend \cite{franciscouzo2025ishihara} from circle-based Ishihara images to polygons, crosses, and stars. Sparse-vertex sampling suffices for circles but leaves gaps in shapes with tips or concavities (\autoref{fig:algo_comparison}, left). To fix this, we overlay a 0.5-pixel grid on the bounding box and classify points via the odd–even ray casting test. By the Jordan curve theorem, odd crossings mark pixels as inside, ensuring complete fillings (\autoref{fig:algo_comparison}, right).

\subsection{Data Statistics}

Our dataset contains 17,100 black-and-white silhouettes and 70,200 camouflaged images at 512$\times$512 resolution. As shown in \autoref{fig:flow_chart_resized}, each camouflaged image is derived from a silhouette with varying palettes, fill geometries, and task types, and is paired with a question–answer. In total, this produces 61 distinct foreground–background configurations and supports nine VQA task types. A subset of 10,800 silhouette–camouflage pairs is reserved for contrastive fine-tuning experiments.

\section{Contrastive Fine-Tuning Framework}
\label{sec:finetune}

Camouflaged images introduce structured clutter: dense distractors and noisy boundaries obscure the global silhouette, leading pretrained VLMs to rely on local textures or spurious cues. Such conditions are rarely represented in large-scale pretraining corpora, leaving models poorly calibrated for figure–ground segregation. To address this gap, we adopt a model-agnostic contrastive fine-tuning framework. Clean silhouettes serve as anchors, their camouflaged counterparts as positives, and unrelated silhouettes as negatives. This design explicitly guides the vision encoder to recover global shape despite camouflage, encouraging models to focus on holistic structure rather than being misled by local background detail.

\subsection{Contrastive Fine-Tuning Method}
\label{subsec:finetune}

Let $\mathcal{B}=\left\{\left(I^O_i, I^C_i\right)\right\}_{i=1}^N$  denote a batch of $N$ image pairs from our training dataset, where $I^O_i$ is the original image and $I^C_i$ is the corresponding camouflaged image. We use the vision encoder $\Phi$ of the VLM to extract patch-level embeddings for each image, followed by average pooling to obtain a single global representation for each image, $\mathbf{x}^O_i = \mathrm{Avg}\left(\Phi\left(I^O_i\right)\right) \text{ and } \mathbf{x}^C_i = \mathrm{Avg}\left(\Phi\left(I^C_i\right)\right)$,
where $\mathbf{x}^O_i$ and $\mathbf{x}^C_i$ are the global embeddings of the silhouette and camouflaged images, respectively.

Similar to \cite{yu2025cafe}, we use the InfoNCE \cite{oord2018representation} loss to align the embeddings of original and camouflaged image pairs.
\begin{equation}
  \mathcal{L}_{con} = -\sum_{i=1}^{\left| \mathcal{B}\right|}\left[\log \frac{h\left(\mathbf{x}^C_i, \mathbf{x}^O_i\right)}{\sum_{j \neq i} h\left(\mathbf{x}^C_i, \mathbf{x}^O_j\right)+h\left(\mathbf{x}^C_i,  \mathbf{x}^O_i\right)}\right] ,
  \label{eq:multiview_loss}
\end{equation}
where $h(\mathbf{x},\mathbf{y})=\exp\left(\frac{\mathbf{x}^\top\mathbf{y}}{\tau}\right)$ is a discriminating function, and $\tau$ is the temperature hyperparameter.

We use embeddings of the silhouette images only for the contrastive learning component, while the camouflaged images are passed through the standard vision-language pipeline to generate answers. This design ensures that only camouflaged images are required during inference. We have the multimodal autoregressive loss $\mathcal{L}_{vlm}\left(\mathbf{y} ; \mathbf{\theta}\right)=-\sum_{i=1}^n log\left[p_\theta\left(y_{i} \mid \Phi\left(I^C_i\right),y_{<i}; \theta\right)\right],$
where $y_{i}$ is the $i$-th token in the response $\mathbf{y}$, and $\theta$ is the trainable parameters. Thus, our total loss is $\mathcal{L}=\alpha \mathcal{L}_{con} + \left(1-\alpha\right) * \mathcal{L}_{vlm},$
where $\alpha$ is the scale parameter.

\begin{figure*}[htbp]
    \centering
    \includegraphics[width=.9\textwidth]{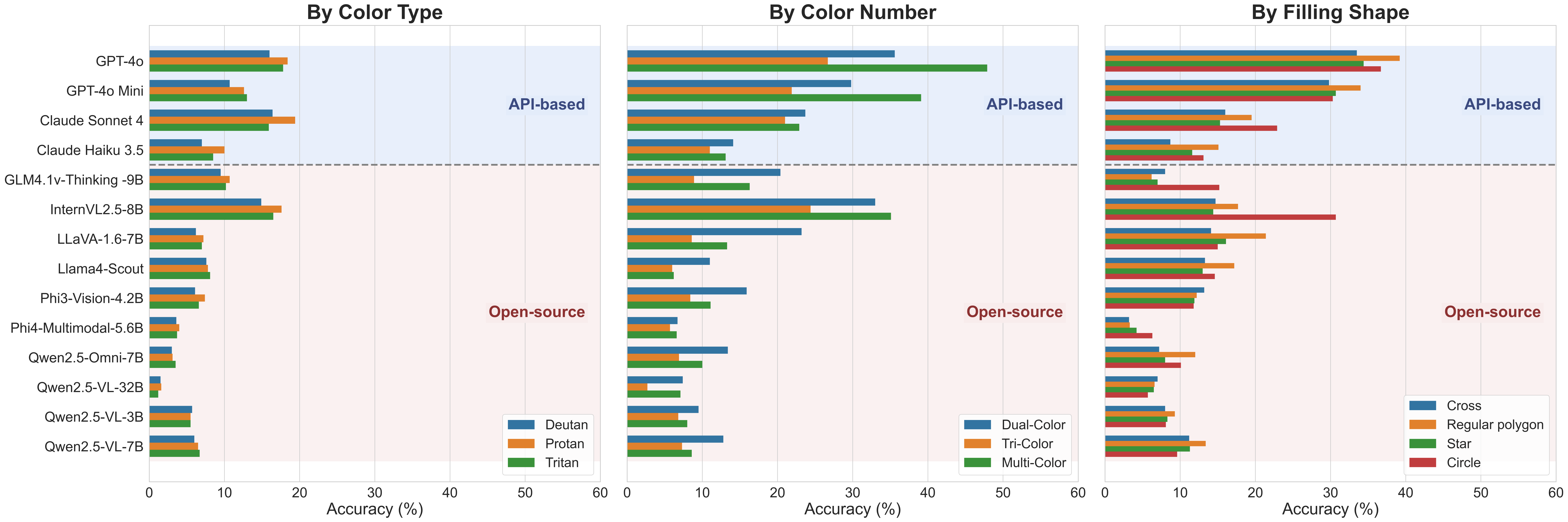}
    \caption{
        \textbf{Evaluation Accuracy Under Different Configurations.}
        Each subplot compares model performance across three types of image configurations:
        (a) CIE color palettes,
        (b) Ishihara color numbers,
        (c) Filling shape types.
    }
    \label{fig:model_comparison_three}
\end{figure*}

\begin{table*}[t]
\centering
\resizebox{\textwidth}{!}{
\begin{tabular}{lccccccccccc}
\toprule
\textbf{} 
& \textbf{Silhouette} 
& \textbf{Math} 
& \textbf{Recognition} 
& \textbf{Occlusion} 
& \textbf{Rotation} 
& \textbf{Count} 
& \textbf{Enumeration} 
& \textbf{Spot\newline Difference} 
& \textbf{Size\newline Comparison} 
& \textbf{Size\newline Sort} 
& \textbf{Overall} \\
\midrule
Humans (average) & 100\% & 87.2\% & 90.2\% & 81.7\% & 82.5\% & 98.8\% & 85.9\% & 86.1\% & 99.2\% & 86.3\% & 88.7\% \\
\midrule
Contra-Qwen ($\alpha=0$) & 51.6\% & 25.1\% & 21.6\% & 15.5\% & 12.0\% & 77.7\% & 4.2\% & 48.3\% & 53.2\% & 27.7\% & 31.7\% 
\\
Contra-Qwen ($\alpha=0.1$)            & 63.8\% & 57.0\% & 46.6\% & 30.0\% & 28.4\% & 85.4\% & 25.2\% & 83.9\% & 85.6\% & 86.1\% & 58.7\% 
\\
Contra-Qwen ($\alpha=0.2$)            & \textbf{72.0\%} & 57.4\% & 47.3\% & 30.3\% & \textbf{30.7\%} & 88.3\% & 32.3\% & 87.1\% & 89.8\% & 85.0\% & 60.9\% 
\\
Contra-Qwen ($\alpha=0.3$)            & 71.4\% & 56.9\% & 45.7\% & 28.8\% & 27.7\% & 87.1\% & 29.7\% & 86.4\% & 88.7\% & 90.1\% & 60.1\% 
\\
Contra-Qwen ($\alpha=0.5$)            & 71.9\% & \textbf{59.6\%} & 47.7\% & 30.0\% & 28.0\% & 87.9\% & 28.7\% & 84.9\% & \textbf{89.9\%} & \textbf{92.7\%} & \textbf{61.0\%} 
\\
Contra-Qwen ($\alpha=0.8$)            & 71.6\% & 54.6\% & \textbf{48.7\%} & \textbf{30.8\%} & 30.4\% & \textbf{89.0\%} & \textbf{33.5\%} & \textbf{87.2\%} & 84.2\% & 87.0\% & 60.6\% 
\\
Contra-Qwen ($\alpha=1$)            & 71.8\% & 53.8\% & 48.1\% & 28.3\% & 28.8\% & 87.8\% & 27.4\% & 85.6\% & 89.3\% & 86.3\% & 59.5\% 
\\
\bottomrule
\end{tabular}
}
\caption{Evaluation results for our finetuned model and ablation studies.}

\label{tab:finetune_result_full}
\end{table*}

\section{Empirical Results}
\label{sec:results_and_analysis}

\subsection{Evaluation Protocol}
We benchmark (i) human participants, (ii) Proprietary models, (iii) open-source models, and (iv) adapted variants trained with our contrastive framework. 
For adaptation, we use three representative backbones: Qwen2.5-VL-7B, InternVL-2.5-8B, and LLaVA-1.6-7B, referred to as Contra-Qwen, Contra-InternVL, and Contra-LLaVA. 
Each model is fine tuned with the vision encoder, projection layers, and language model set as trainable, using $\alpha=0.5$, $\tau=0.7$, and 10{,}800 silhouette–camouflage pairs where silhouettes serve as anchors, camouflaged renderings as positives, and unrelated silhouettes as negatives. 
Training is performed on 8 A100 GPUs with batch size 16 using the Adam optimizer. Learning rates are $1\times10^{-7}$ for the language model, $1\times10^{-5}$ for the projection layers, and $1\times10^{-6}$ for the vision encoder. Training for 2 epochs.

We use GPT-4o-mini to evaluate model outputs for open-ended tasks, followed by human validation. Ten human evaluators also answer the same set of test images. For the leaderboard, the Silhouette column reports performance on clean silhouettes and is not included in the overall average.

\subsection{Results and Analysis}
As shown in \autoref{tab:recognition_accuracy_detailed_weighted}, recognition accuracy is reported for Silhouette source images and nine reasoning tasks. The overall column averages the camouflage tasks only.

\noindent \textbf{Humans vs. VLMs.} Human participants achieve much higher accuracy across all tasks, while evaluated models fall significantly short. On camouflaged image tasks, VLMs blur hue boundaries, mis-segment foregrounds, and fail to integrate dispersed dots, as shown in \autoref{tab:recognition_accuracy_detailed_weighted}. They are ``shape-smart'' in clean silhouettes but ``pattern-blind'' in textured fields, exposing limits in figure–ground segregation. Without objectives that preserve fine chromatic edges and promote global grouping, VLMs remain below human level.

\noindent \textbf{Baselines.} GPT-4o leads with 20.4\% overall accuracy, followed by InternVL2.5-8B at 18.6\%. Qwen2.5-VL-32B performs worst at 2.1\%. InternVL2.5-8B shows relative strength on spot-the-difference and size comparison.

\noindent \textbf{Impact of color, shape, and task type.} \autoref{fig:model_comparison_three} shows that accuracy is relatively stable across the sRGB-bounded palette configurations but drops with Tri-color Ishihara settings. Cross-shaped fills perform worst due to fragmented geometry. Accuracy varies as tasks differ in cognitive and perceptual demands. Size Comparison and Spot Difference are easier since they rely on visual contrast and local comparisons, which VLMs handle with spatial attention and pattern matching. In contrast, Rotation, Enumeration, and Math require abstract reasoning, spatial transformations, and symbolic understanding of global structures under noise, which require deeper semantic and geometric comprehension.

\noindent \textbf{VLMs excel on silhouettes but fail on camouflaged images.} \autoref{tab:recognition_accuracy_detailed_weighted} shows they perform well on clean, high-contrast silhouette images but struggle with figure–ground segregation in textured or colorful backgrounds. Unlike humans, they lack strong global grouping and under-utilize color/context, revealing limits in visual abstraction and the need for models that better disentangle objects from background.

\noindent \textbf{Larger models are not always better.} \autoref{tab:recognition_accuracy_detailed_weighted} shows that size alone does not ensure higher accuracy. GPT-4o leads, but large open-source models like Qwen2.5-VL-32B trail smaller ones. Low-resolution, large-patch encoders in big models often blur the faint hue edges in the patterns, while smaller models fine-tuned at higher resolution can preserve them. Adding language parameters without widening the vision–language bottleneck forces noisy feature compression, and because camouflaged images are rare in pretraining data, extra parameters often reinforce spurious correlations rather than learning the abstraction needed for figure–ground segregation.

\noindent \textbf{Contrastive Adaptation.} 
Contra-Qwen reaches 61.0\% overall, increasing by 54.1\% over its baseline, with gains of 58.5\% on math and 86.7\% on size sorting. 
Contra-InternVL rises to 60.6\%, increasing by 42.0\% overall, with counting improved by 62.4\% and size comparison by 23.8\%. 
Contra-LLaVA improves to 56.6\%, increasing by 48.5\% overall, with occlusion improved by 28.4\% and spot-the-difference by 55.3\%. 
On silhouettes, all adapted models exceed 65\%, with Contra-Qwen at 71.9\%, rivaling much larger proprietary systems. 
These consistent gains across distinct backbones show the framework is model-agnostic and effective for recovering global shape under camouflage.

\subsection{Ablation Studies}
We assess the effect of the contrastive loss weight $\alpha$ using the best-performing model Contra-Qwen. Values range from 0 to 1. At $\alpha=0$ the model reduces to the baseline Qwen without contrastive loss, while $\alpha=1$ corresponds to contrastive loss only. Results in \autoref{tab:finetune_result_full} show that moderate values yield the best trade-off. Performance improves steadily as $\alpha$ increases from 0 to 0.2, peaking at $\alpha=0.5$ with an overall accuracy of 61.0\%. Larger $\alpha$ values further boost recognition and occlusion but reduce reasoning performance, leading to lower overall scores.

\section{Conclusion}
\label{sec:conclusion}
We introduce ChromouVQA, a large-scale benchmark for evaluating the perceptual and reasoning abilities of VLMs on camouflaged images. Our results show a substantial gap between human and model performance in tasks involving abstract or visually ambiguous patterns, highlighting current limitations. To address this, we propose a model-agnostic contrastive learning framework that significantly improves performance. Future work will expand the benchmark with more diverse patterns and task types to further challenge VLMs.

\bibliographystyle{IEEEbib}
\bibliography{main}

\end{document}